\documentclass[conference]{IEEEtran}
\IEEEoverridecommandlockouts
\usepackage{amsmath}
\usepackage{graphicx}
\usepackage{hyperref}
\usepackage{tikz}
 \usepackage{placeins} 
 \usepackage{array}
  \usepackage{caption}
  \usepackage{cite}
  \usepackage{ragged2e}
\usepackage{makecell} 
  \usepackage[caption=false,font=footnotesize]{subfig} 
\usepackage{tikz}

\usetikzlibrary{positioning}   
\usetikzlibrary{arrows.meta}   
\usetikzlibrary{fit}    

 \makeatletter
    \def\ps@IEEEtitlepagestyle{
      \def\@oddfoot{\mycopyrightnotice}
      \def\@evenfoot{}
    }
    \def\mycopyrightnotice{
      {\footnotesize 979-8-3315-7867-1/25/\$31.00~\copyright~2025 IEEE\hfill} 
     \gdef\mycopyrightnotice{}
   }

    \usepackage[mathlines,switch]{lineno}

     \makeatletter
    
    \let\old@ps@IEEEtitlepagestyle\ps@IEEEtitlepagestyle
    \def\confheader#1{%
        \def\ps@IEEEtitlepagestyle{%
            \old@ps@IEEEtitlepagestyle%
            \def\@oddhead{\strut\hfill#1\hfill\strut}%
            \def\@evenhead{\strut\hfill#1\hfill\strut}%
        }%
        \ps@headings%
    }
    \makeatother
    
    \confheader{%
            \parbox{20cm}{2025 28th International Conference on Computer and Information Technology (ICCIT)\\19-21 December, Cox’s Bazar, Bangladesh}
    }

\begin{document}

\title{\LARGE Detecting AI-Generated Paraphrases in Bengali: A Comparative Study of Zero-Shot and Fine-Tuned Transformers}

\author{
\IEEEauthorblockN{Md. Rakibul Islam\IEEEauthorrefmark{1}, Most. Sharmin Sultana Samu\IEEEauthorrefmark{2}, Md. Zahid Hossain\IEEEauthorrefmark{3}, Farhad Uz Zaman\\ \IEEEauthorrefmark{4}, Md. Kamrozzaman Bhuiyan\IEEEauthorrefmark{5}}
\IEEEauthorblockA{
\IEEEauthorrefmark{1} \IEEEauthorrefmark{3}Department of CSE, Ahsanullah University of Science and Technology, Bangladesh.\\
\IEEEauthorrefmark{2}Department of CSE, BRAC University, Bangladesh.\\
\IEEEauthorrefmark{4}Department of CSE, Southeast University, Bangladesh.\\
\IEEEauthorrefmark{5}Enosis Solutions, Bangladesh.
}
\IEEEauthorblockA{
Email: rakib.aust41@gmail.com\IEEEauthorrefmark{1}, \IEEEauthorrefmark{2}sharminsamu130@gmail.com, \IEEEauthorrefmark{3}zahidd16@gmail.com,\\farhad.zaman@seu.edu.bd
\IEEEauthorrefmark{4}, \IEEEauthorrefmark{5}kamrozzamaan@gmail.com
}
}

\twocolumn[
\begin{@twocolumnfalse}
\vfill

\fontsize{20}{24}\selectfont

\textbf{IEEE Copyright Notice}

\vspace{1em}

\fontsize{14}{17}\selectfont   

\noindent
\begin{minipage}{1.0\textwidth}
\justifying

\textcopyright\ 2025 IEEE. Personal use of this material is permitted. Permission from IEEE must be obtained for all other uses, in any current or future media, including reprinting/republishing this material for advertising or promotional purposes, creating new collective works, for resale or redistribution to servers or lists, or reuse of any copyrighted component of this work in other works. \\

\vspace{2em}

This work has been accepted for publication in \textbf{2025 28th International Conference on Computer and Information Technology (ICCIT)}. The final published version will be available via IEEE Xplore. \\

DOI: \textit{TBD}

\end{minipage}

\vfill
\end{@twocolumnfalse}
]

\maketitle
\begin{abstract}
Large language models (LLMs) can produce text that closely resembles human writing. This capability raises concerns about misuse, including disinformation and content manipulation. Detecting AI-generated text is essential to maintain authenticity and prevent malicious applications. Existing research has addressed detection in multiple languages, but the Bengali language remains largely unexplored. Bengali’s rich vocabulary and complex structure make distinguishing human-written and AI-generated text particularly challenging. This study investigates five transformer-based models: XLM-RoBERTa-Large, mDeBERTaV3-Base, BanglaBERT-Base, IndicBERT-Base and MultilingualBERT-Base. Zero-shot evaluation shows that all models perform near chance levels (around 50\% accuracy) and highlight the need for task-specific fine-tuning. Fine-tuning significantly improves performance, with XLM-RoBERTa, mDeBERTa and MultilingualBERT achieving around 91\% on both accuracy and F1-score. IndicBERT demonstrates comparatively weaker performance, indicating limited effectiveness in fine-tuning for this task. This work advances AI-generated text detection in Bengali and establishes a foundation for building robust systems to counter AI-generated content.\\

\renewcommand
\IEEEkeywordsname{Keywords}

\begin{IEEEkeywords}
AI-paraphrased Text Detection, Bengali Text Classification, Human vs AI, Large Language Models, Natural Language Processing
\end{IEEEkeywords}

\end{abstract}

\section{Introduction}
Recent advances in large language models (LLMs) have enabled machines to generate text that closely resembles human writing, powering tasks such as summarization, translation, paraphrasing, creative writing and code generation. While models like ChatGPT and GPT-4 \cite{openai2023gpt} showcase impressive capabilities, they also raise concerns over misuse for misinformation, propaganda and large-scale content manipulation. Public APIs make these tools widely accessible, heightening risks and underscoring the need for reliable detection methods. Misleading websites hosting AI-written articles highlight the urgency of robust systems to identify machine-generated text. This study addresses this challenge by focusing on detecting AI-generated news and paraphrased content in Bengali.

Detecting AI-generated content is essential to maintain information integrity and trust in digital communications. Prior research has explored differentiating human-written and AI-generated text in English. A research gap persists in the Bengali language due to its complex grammar, extensive vocabulary and large set of letters. These linguistic characteristics make distinguishing between human and machine-generated Bengali text particularly challenging. \cite{11022427} proposed a hybrid BiLSTM-SVM model to classify Bengali text as either human-written or ChatGPT-paraphrased, achieving an accuracy of 82.83\%.

In this work, we address the question: “Can transformer-based models surpass traditional deep learning in detecting human-written versus ChatGPT-paraphrased Bengali text and which model shows the greatest improvement?” We investigate transformer-based models for classifying human-written and ChatGPT-paraphrased Bengali text. The BanglaTextDistinguish dataset \cite{11022427} is used, comprising 6,644 instances from newspapers, social media and textbooks, with AI-generated paraphrases produced using GPT-3.5. Five transformer-based models, including XLM-RoBERTa-Large \cite{conneau-etal-2020-unsupervised}, mDeBERTaV3-Base \cite{he2023debertav3improvingdebertausing}, BanglaBERT-Base \cite{Sagor_2020}, IndicBERT-Base \cite{kakwani2020indicnlpsuite} and MultilingualBERT-Base \cite{DBLP:journals/corr/abs-1810-04805}, are assessed in zero-shot settings and fine-tuned for the classification task. We present a comparative analysis of our transformer models against recent deep learning and machine learning approaches. The study aims to determine the most effective model and enhance AI-generated text detection for the Bengali language.

Our key contributions are as follows:
\begin{itemize}
    \item We present the first study using transformer-based models to distinguish between human-written and ChatGPT-paraphrased Bengali text.
    \item We evaluate five transformer architectures (XLM-RoBERTa, mDeBERTa, BanglaBERT, IndicBERT and MultilingualBERT) on the BanglaTextDistinguish dataset of 6,644 diverse instances, in both zero-shot and fine-tuned settings.
    \item We compare transformer models with existing BiLSTM-SVM and deep learning baselines, showing significant improvements in accuracy and robustness for detecting Bengali AI-generated text.
    \item We emphasize the potential of fine-tuned transformer models for developing robust detection systems in low-resource languages.
\end{itemize}

The paper is structured as follows. Section II reviews existing works on AI generated text detection and related studies in text processing. Section III provides background study relevant to this research. Section IV describes the proposed methodology, including zero-shot classification and fine-tuning strategies with different models. Section V presents the experimental results with detailed performance analysis. Section VI concludes the paper and highlights possible directions for future research in Bengali AI generated text detection.

\section{Related Works}
Recent studies have explored AI-generated text detection using transformers, ensemble models and information-theoretic methods. Domain adaptation, multilingual setups and low-resource approaches such as contrastive learning, residual subspace methods and hybrid deep learning classifiers have also shown strong results in English and other languages. However, challenges remain with paraphrasing, domain shifts, dataset limitations and cross-lingual generalization.

\cite{chakraborty2023possibilitiesaigeneratedtextdetection} used information-theoretic methods with GPT-2, GPT-3.5-Turbo, LLaMA, oBERTa and Logistic Regression, achieving 97\% AUROC. \cite{wang2024aigeneratedtextdetectionclassification} applied a BERT-based model to 1,378 texts, reaching 97.71\% accuracy, while \cite{wang-etal-2023-seqxgpt} proposed SeqXGPT with log probabilities from GPT2-xl, GPT-J and LLaMA, reporting Macro-F1 above 95\%. Ensemble methods in \cite{10594194} and \cite{abburi-etal-2023-simple} achieved ROC-AUC up to 0.975 and Fmacro 97.9\% and \cite{lai2024adaptive} improved detection accuracy from 62.9\% to 72.5\% using adaptive ensembles. \cite{yadagiri-etal-2024-detecting} combined linguistic and statistical features with transformers, reaching 99.73\% accuracy on HC3-English.

\cite{bhattacharjee-etal-2023-conda} introduced ConDA with RoBERTa for contrastive domain adaptation, improving performance by 31.7\%, while \cite{kuznetsov-etal-2024-robust} used residual subspaces for cross-domain robustness gains of 14\%. \cite{bhattacharjee2024eagledomaingeneralizationframework} proposed EAGLE with domain adversarial and contrastive learning, showing strong generalization on TuringBench. A hybrid POS-tagged BiLSTM in \cite{10971184} reached 88\% accuracy and \cite{Schaaff2024} reported F1-scores up to 99\% using XGBoost, Random Forest and MLP across four languages.

For Bengali, \cite{11022427} developed a BiLSTM-SVM achieving 82.83\% accuracy. \cite{ALGHAMDI2025100353} proposed ABERT with 99.09\% accuracy and reduced parameters, while \cite{LAU2025100151} showed human paraphrasing increased TPR but reduced AUROC. \cite{jawaid-etal-2024-human} combined DistilBERT with post-processing, achieving 87.5\% accuracy on SemEval-2024 and \cite{10994761} used AraELECTRA and AraBERT for dialectal Arabic, finding rephrased text particularly challenging.

\section{Background Study}

\subsection{Transformer Models for AI-Paraphrased Text Classification}
We explore five transformer-based models for detecting AI-paraphrased Bengali text. XLM-RoBERTa-Large \cite{conneau-etal-2020-unsupervised} is a multilingual model trained on 100 languages. It is designed for strong cross-lingual transfer. mDeBERTaV3-Base \cite{he2023debertav3improvingdebertausing} improves on DeBERTa with better attention mechanisms and training efficiency. It performs well across languages. BanglaBERT-Base \cite{Sagor_2020} is a monolingual model trained on large Bengali corpora, making it effective for Bengali tasks. IndicBERT-Base \cite{kakwani2020indicnlpsuite} is a multilingual model optimized for 12 Indian languages, including Bengali, which makes it efficient in low-resource settings. MultilingualBERT-Base \cite{DBLP:journals/corr/abs-1810-04805} is an early multilingual model trained on 104 languages, serving as a strong baseline for multilingual text classification. These models offer complementary strengths for AI-generated text detection, including cross-lingual transfer, language-specific representation and resource-efficient training.

\subsection{Evaluation Metrics}\label{A9}
Accuracy represents the percentage of correctly classified samples among all samples and indicates overall classification performance. Precision measures the proportion of true positive predictions among all predicted positives and reflects the reliability of positive detections. Recall calculates the proportion of true positives identified among all actual positives and indicates the sensitivity of the model. F1 Score (Binary) is the harmonic mean of precision and recall in binary classification and balances false positives and false negatives. F1 Score (Macro) averages F1 scores across all classes equally and reflects balanced performance in multi-class settings. AUROC measures the ability of the model to distinguish between classes across thresholds and indicates overall discrimination capability. A larger AUROC shows stronger separability, while a smaller AUROC shows weaker performance. Brier Score measures the accuracy of probabilistic predictions and indicates calibration quality. A smaller Brier Score reflects better calibration and reliability, while a larger score reflects poor calibration.

\section{Methodology}
\subsection{Dataset and Preprocessing}
The BanglaTextDistinguish \cite{11022427} dataset is employed in this study, which contains Bengali texts sourced from newspapers, textbooks and social media. Newspaper and textbook sentences represent formal writing, while social media provides informal usage, creating a balanced mix of linguistic styles. A total of 3322 sentences were collected and paraphrased with GPT-3.5, resulting in 6644 samples. Sentence lengths follow a normal distribution with an average of 43.54 words and a standard deviation of 23.17. Duplicate entries are removed to eliminate redundancy. Null instances are deleted to ensure data integrity, leaving 6640 valid samples. Class labels are transformed into integers through label encoding for computational processing. The dataset is divided into training, validation and testing sets at a 60:20:20 ratio to support unbiased evaluation. Each subset is reformatted according to the input requirements of transformer models, including tokenization, attention mask generation and special token insertion, ensuring compatibility with both multilingual and Bengali-specific architectures.

\subsection{Zero-Shot Inference Models}
\begin{figure}[hbt!] 
    \includegraphics[width=80mm,scale=0.65]{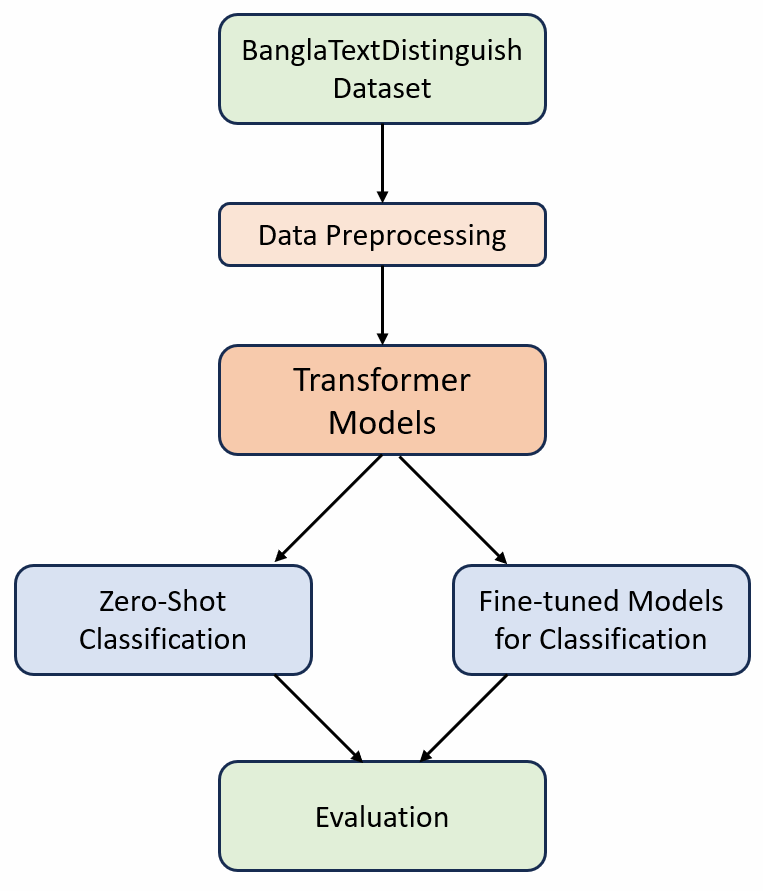}
    \caption{Proposed Methodology}
    \label{fig:gr1}
\end{figure}

The zero-shot experiments use the BanglaTextDistinguish \cite{11022427} dataset containing human-written and GPT-3.5 paraphrased sentences. Each input sentence is tokenized according to model requirements, with special tokens, attention masks, truncation and padding applied to match the maximum sequence length. The XLM-RoBERTa-Large-xnli and mDeBERTaV3-Base-xnli models are loaded via the Hugging Face pipeline for zero-shot classification. BanglaBERT-Base, IndicBERT-Base and MultilingualBERT-Base models are used to generate contextual embeddings by mean pooling the last hidden state for each token. Label embeddings are precomputed from Bengali descriptions of “Human-written” and “AI-generated” classes. Cosine similarity between text embeddings and label embeddings determines the predicted label, while the AI-generated probability is recorded. The models return predicted labels, similarity scores and class probabilities. Figure \ref{fig:gr1} shows the proposed methodology of our work.

\subsection{Fine-Tuned Models}
The fine-tuned models are trained on the BanglaTextDistinguish \cite{11022427} dataset comprising human-written sentences labeled as 0 and GPT-3.5 paraphrased texts labeled as 1. Tokenization is applied using model-specific tokenizers including XLM-RoBERTa-Large, mDeBERTaV3-Base, BanglaBERT-Base, IndicBERT-Base and MultilingualBERT-Base, with truncation, padding and maximum sequence lengths ranging from 128 to 256 tokens. Encoded inputs include input IDs, attention masks and label tensors. The tokenized data is wrapped in custom PyTorch Dataset classes for efficient loading during training and evaluation. Sequence classification models are initialized with two output labels and trained using the Hugging Face Trainer API. Training uses cross-entropy loss optimized via AdamW and mixed precision (fp16) is enabled on GPUs when available. Early stopping monitors validation metrics with patience of 2–3 epochs and the best model is loaded at the end of training. Evaluation converts logits to probabilities via sigmoid and predicts the class with the highest probability.

\subsection{Experimental Setup}
The experiments were conducted using an NVIDIA RTX 3060 GPU. Standard deep learning libraries including PyTorch and Hugging Face Transformers were employed. Stratified sampling splits the data into training, validation and test sets in a 60:20:20 ratio to maintain class balance. Hyperparameters vary across models, including learning rates from 1e-5 to 2e-5, batch sizes of 8–16, weight decay 0.01, gradient accumulation of 1–2 and 2–9 training epochs. Hyperparameter sweeps are performed to identify optimal settings for each model. Table \ref{tab:training_params} reports the training hyperparameters used for different models.

\begin{table}[!ht]
\centering
\caption{\label{tab:training_params} Training hyperparameters for fine-tuned Bengali AI text detection models. Learning Rate and Batch Size are abbreviated as LR and BS respectively.}
\begin{tabular}{|l|c|c|c|c|c|}
\hline
\textbf{Model Name} & \textbf{\makecell{LR}} & \textbf{\makecell{BS}} & \textbf{\makecell{Weight\\ Decay}} & \textbf{Epochs} & \textbf{\makecell{Seq\\Length}} \\
\hline
XLM-RoBERTa-Large & 1e-5 & 8 & 0.01 & 5 & 256 \\
mDeBERTaV3-Base & 2e-5 & 16 & 0.01 & 4 & 256 \\
BanglaBERT-Base & 1e-5 & 16 & 0.01 & 3 & 128 \\
IndicBERT-Base & 2e-5 & 16 & 0.01 & 9 & 256 \\
MultilingualBERT-Base & 2e-5 & 16 & 0.01 & 2 & 256 \\
\hline
\end{tabular}
\end{table}

\section{Result Analysis}
In this section, we present the experimental results of both zero-shot inference models and fine-tuned models.
\subsection{Zero-Shot Classification}
Zero-shot transformer models showed limited effectiveness in detecting AI-generated Bengali paraphrases.

\begin{table}[!ht]
\centering
\caption{\label{tab:zeroshot_model_metrics_bengali} Performance comparison of zero-shot transformer models for Bengali AI-paraphrased text detection. Acc, Prec, Rec and F1 stand for Accuracy, Precision, Recall and F1 Score (Binary) respectively. Values are presented in percentage.}
\begin{tabular}{|l|c|c|c|c|}
\hline
\textbf{Model Name} & \textbf{Acc} & \textbf{Prec} & \textbf{Rec} & \textbf{F1} \\
\hline
\makecell{XLM-RoBERTa-Large-xnli} & 49.02 & 47.71 & 22.02 & 30.13 \\
\makecell{mDeBERTaV3-Base-xnli} & 49.17 & 43.75 & 6.33 & 11.07 \\
\makecell{BanglaBERT-Base} & 50.04 & 50.00 & 99.55 & 66.57 \\
\makecell{IndicBERT-Base} & \textbf{50.34} & \textbf{50.16} & 92.47 & 65.04 \\
\makecell{MultilingualBERT-Base} & 50.26 & 50.11 & \textbf{99.70} & \textbf{66.70} \\
\hline
\end{tabular}
\end{table}

BanglaBERT-Base and MultilingualBERT-Base achieved around 50\% accuracy with very high recall (99.55\% and 99.70\%) but only moderate precision, while IndicBERT-Base reached 50.34\% accuracy and 92.47\% recall, indicating some misclassification of human text. XLM-RoBERTa-Large-xnli and mDeBERTaV3-Base-xnli performed worst, with accuracy below 50\% and recall as low as 6.33\%. F1 scores were low for these two models but higher (65–67\%) for BanglaBERT-Base, IndicBERT-Base and MultilingualBERT-Base. Overall, zero-shot models can detect some AI-generated content but lack task-specific training for reliable classification. Table \ref{tab:zeroshot_model_metrics_bengali} summarizes the performance metrics.

XLM-RoBERTa-Large-xnli achieved a macro F1 of 45.00\%, AUROC 49.03\% and Brier 29.17\%, indicating moderate discrimination but weak calibration. mDeBERTaV3-Base-xnli performed worst (F1 37.74\%, AUROC 45.61\%), while BanglaBERT-Base had lower F1 (33.88\%) but the best calibration (Brier 25.26\%). IndicBERT-Base balanced classes better (F1 39.66\%) but with poor calibration (Brier 37.83\%) and MultilingualBERT-Base recorded the highest AUROC (58.06\%) with F1 34.24\%. Zero-shot models lacked robustness, showing trade-offs between discrimination, balance and calibration. Table \ref{tab:zeroshot_model_metrics_bengali_additional} presents these metrics.

\begin{table}[!ht]
\centering
\caption{\label{tab:zeroshot_model_metrics_bengali_additional} Performance comparison of zero-shot transformer models for Bengali AI-paraphrased text detection. F1 (Macro), AUROC and Brier Score values are presented in percentage.}
\begin{tabular}{|l|c|c|c|}
\hline
\textbf{Model Name} & \textbf{F1 (Macro)} & \textbf{AUROC} & \textbf{Brier Score} \\
\hline
\makecell{XLM-RoBERTa-Large-xnli} & \textbf{45.00} & 49.03 & 29.17 \\
\makecell{mDeBERTaV3-Base-xnli} & 37.74 & 45.61 & 32.24 \\
\makecell{BanglaBERT-Base} & 33.88 & 55.74 & \textbf{25.26} \\
\makecell{IndicBERT-Base} & 39.66 & 46.22 & 37.83 \\
\makecell{MultilingualBERT-Base} & 34.24 & \textbf{58.06} & 27.60 \\
\hline
\end{tabular}
\end{table}

\subsection{Fine-Tuned Classification}
The fine-tuned transformer models exhibited significantly improved performance in detecting AI-generated Bengali paraphrases compared to zero-shot models. 

\begin{table}[!ht]
\centering
\caption{\label{tab:fine_tuned_model_metrics} Performance comparison of fine-tuned transformer models for Bengali AI-paraphrased text detection. Values are presented in percentage.}
\begin{tabular}{|l|c|c|c|c|}
\hline
\textbf{Model Name} & \textbf{Acc} & \textbf{Prec} & \textbf{Rec} & \textbf{F1} \\
\hline
\makecell{XLM-RoBERTa-Large} & \textbf{91.50} & \textbf{95.84} & 86.75 & \textbf{91.07} \\
\makecell{mDeBERTaV3-Base} & 91.35 & 94.06 & 88.25 & 91.06 \\
\makecell{BanglaBERT-Base} & 88.26 & 90.84 & 85.09 & 87.87 \\
\makecell{IndicBERT-Base} & 74.25 & 78.75 & 66.42 & 72.06 \\
\makecell{MultilingualBERT-Base} & 90.82 & 90.69 & \textbf{90.96} & 90.83 \\
\hline
\end{tabular}
\end{table}

XLM-RoBERTa-Large achieved the highest accuracy of 91.50\% and precision of 95.84\%, with an F1 score of 91.07\%, demonstrating robust identification of AI-generated text while maintaining low false positives. mDeBERTaV3-Base closely followed with an accuracy of 91.35\%, precision of 94.06\% and F1 score of 91.06\%, reflecting balanced and reliable classification. MultilingualBERT-Base showed slightly lower accuracy at 90.82\% but achieved the highest recall of 90.96\%, indicating strong detection of positive instances. BanglaBERT-Base and IndicBERT-Base displayed moderate performance, with accuracies of 88.26\% and 74.25\% respectively, highlighting some limitations in identifying AI-generated paraphrases. Overall, the table underscores that fine-tuning transformer models substantially enhances classification performance on the BanglaTextDistinguish dataset. Table \ref{tab:fine_tuned_model_metrics} summarizes the detailed performance metrics.

XLM-RoBERTa-Large led the results with an F1 (Macro) score of 91.48\%, AUROC of 96.87\% and a low Brier Score of 8.03\%, reflecting both high classification accuracy and well-calibrated probability estimates. mDeBERTaV3-Base followed closely with an F1 (Macro) of 91.34\%, AUROC of 96.39\% and Brier Score of 7.72\%. MultilingualBERT-Base also performed robustly, achieving an F1 (Macro) of 90.82\%, AUROC of 96.77\% and the lowest Brier Score of 7.06\%, indicating excellent probability calibration. BanglaBERT-Base showed slightly lower performance (F1: 88.25\%, AUROC: 94.53\%, Brier Score: 10.28\%), whereas IndicBERT-Base lagged behind with an F1 (Macro) of 74.09\%, AUROC of 79.67\% and the highest Brier Score of 20.77\%, suggesting comparatively weaker detection capability and calibration. Overall, fine-tuning on the Bengali AI-paraphrased text dataset significantly improved classification performance, discrimination ability and probability calibration over zero-shot approaches. Table \ref{tab:fine_tuned_model_metrics_additional} presents the detailed metrics for all fine-tuned models.

\begin{table}[!ht]
\centering
\caption{\label{tab:fine_tuned_model_metrics_additional} Performance comparison of fine-tuned transformer models for Bengali AI-paraphrased text detection. F1 (Macro), AUROC and Brier Score values are presented in percentage.}
\begin{tabular}{|l|c|c|c|}
\hline
\textbf{Model Name} & \textbf{F1 (Macro)} & \textbf{AUROC} & \textbf{Brier Score} \\
\hline
\makecell{XLM-RoBERTa-Large} & \textbf{91.48} & \textbf{96.87} & 8.03 \\
\makecell{mDeBERTaV3-Base} & 91.34 & 96.39 & 7.72 \\
\makecell{BanglaBERT-Base} & 88.25 & 94.53 & 10.28 \\
\makecell{IndicBERT-Base} & 74.09 & 79.67 & 20.77 \\
\makecell{MultilingualBERT-Base} & 90.82 & 96.77 & \textbf{7.06} \\
\hline
\end{tabular}
\end{table}

\begin{figure}[hbt!] 
    \includegraphics[width=90mm,scale=0.85]{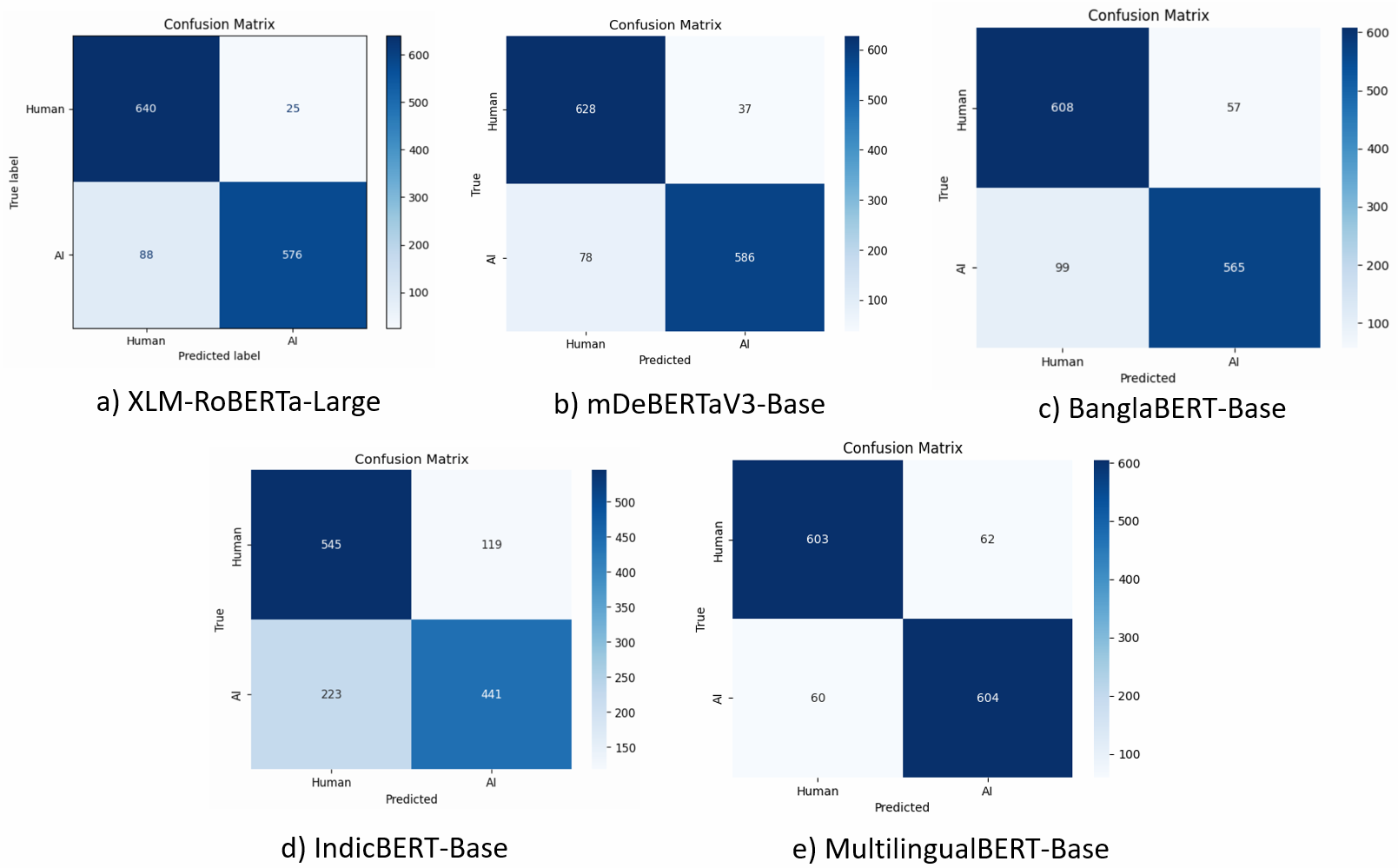}
    \caption{Confusion matrices of the fine-tuned classification models.}
    \label{fig:gr2}
\end{figure}

Figure \ref{fig:gr2} presents the confusion matrices of five fine-tuned models. XLM-RoBERTa-Large achieves strong performance with very low misclassification for both human and AI texts. mDeBERTaV3-Base also performs well, with slightly more errors than XLM-RoBERTa-Large but still balanced across both classes. BanglaBERT-Base shows a moderate increase in misclassification, with higher confusion for AI texts compared to human. IndicBERT-Base performs relatively poorly, with substantial misclassification for both human and AI texts, indicating weaker discriminative ability. MultilingualBERT-Base achieves robust performance, with low errors in both classes and the most balanced results across human and AI detection.

\begin{figure}[hbt!] 
    \includegraphics[width=90mm,scale=0.85]{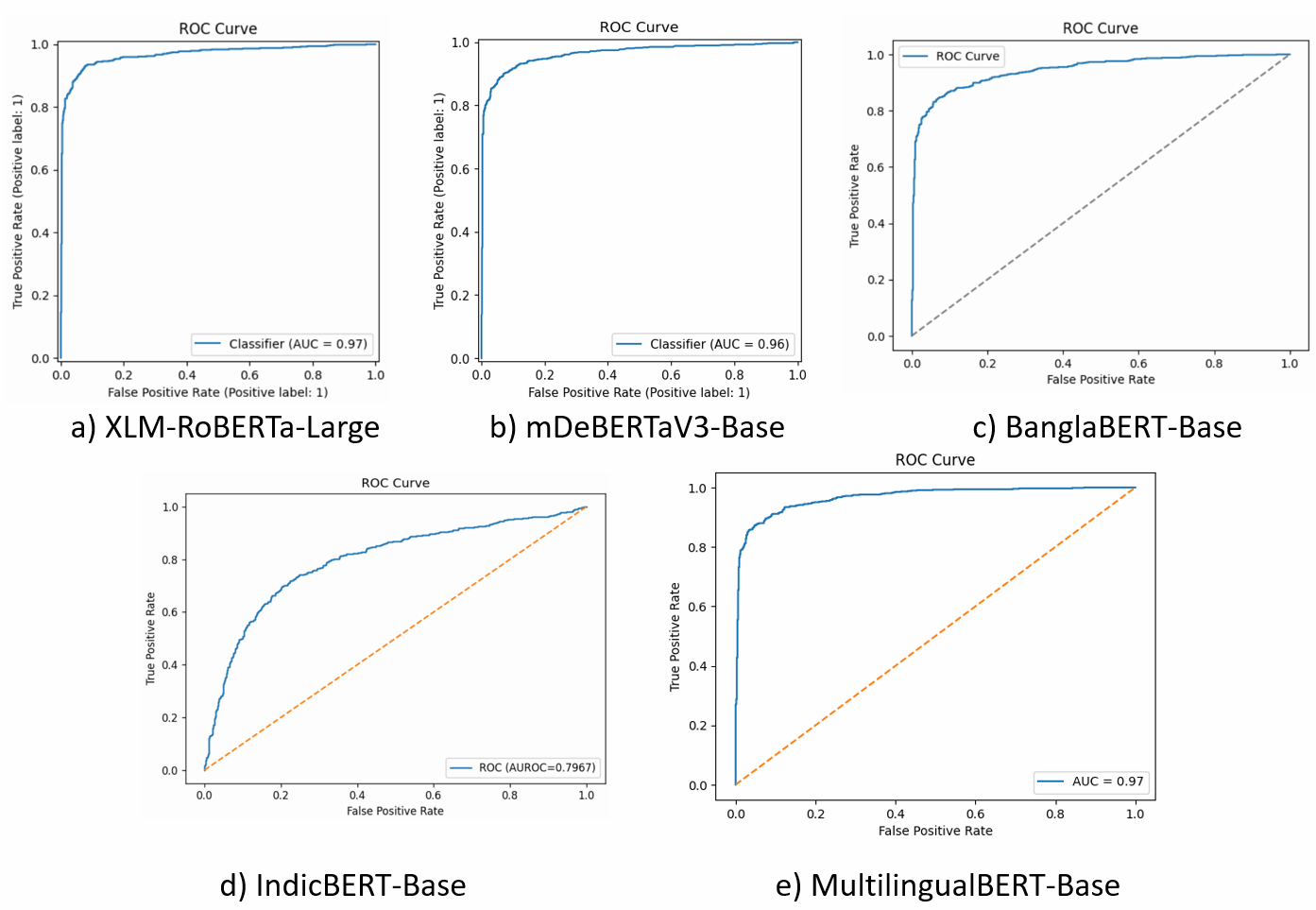}
    \caption{ROC curve of the fine-tuned classification models.}
    \label{fig:gr3}
\end{figure}

Figure \ref{fig:gr3} presents the ROC curves of five fine-tuned models. XLM-RoBERTa-Large demonstrates excellent separability with an AUC of 96.87\%, and the curve rises sharply toward the top left corner, indicating strong classification ability. mDeBERTaV3-Base also performs strongly with an AUC of 96.39\%, showing a similarly steep rise and reliable discrimination between classes. BanglaBERT-Base achieves robust performance with an AUC close to 94.53\%, though the curve is slightly less steep than XLM-RoBERTa-Large and mDeBERTaV3-Base. IndicBERT-Base shows weaker performance with an AUC of about 79.67\%, where the curve is closer to the diagonal and rises more gradually, reflecting moderate classification strength. MultilingualBERT-Base achieves strong results with an AUC of 96.77\%, matching XLM-RoBERTa-Large and exhibiting one of the steepest curves with near-ideal separability.

\begin{figure}[hbt!] 
    \includegraphics[width=90mm,scale=0.85]{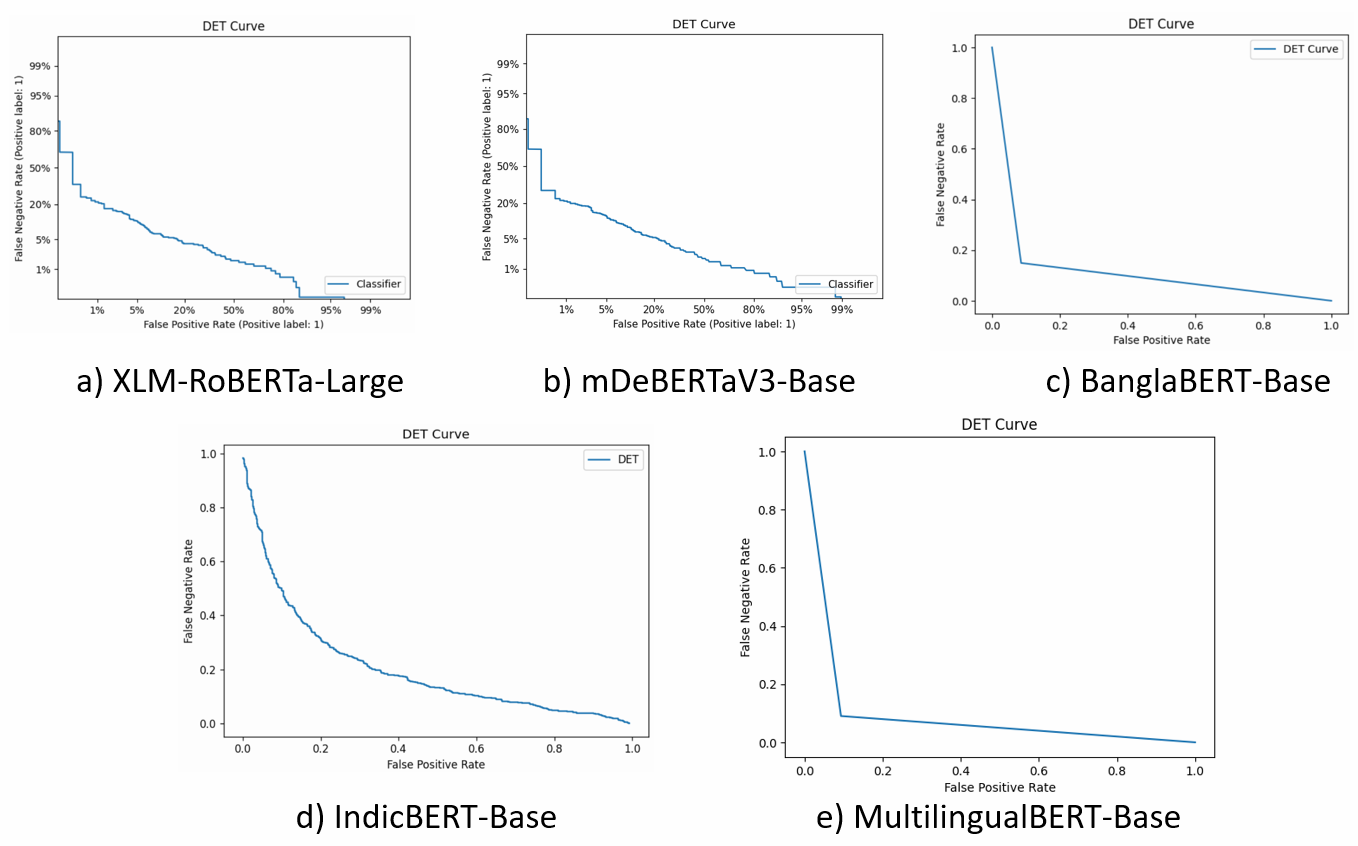}
    \caption{DET curve of the fine-tuned classification models.}
    \label{fig:gr4}
\end{figure}

The Detection Error Tradeoff (DET) curves in Figure \ref{fig:gr4} show the error tradeoff across five fine-tuned models. XLM-RoBERTa-Large demonstrates strong performance with a smooth downward-sloping curve, indicating low error rates at optimal thresholds. mDeBERTaV3-Base shows a similar pattern with slightly higher error rates than XLM-RoBERTa-Large, but still maintains a balanced tradeoff between false positives and false negatives. BanglaBERT-Base presents a very steep curve with minimal error at optimal thresholds, suggesting strong separability but less smoothness in error decline. IndicBERT-Base performs comparatively weaker, with a slower decline and higher error rates across most thresholds, reflecting limited discriminative ability. MultilingualBERT-Base achieves robust results with a sharp curve and very low error rates at optimal thresholds, comparable to XLM-RoBERTa-Large.

\begin{figure}[hbt!] 
    \includegraphics[width=90mm,scale=0.85]{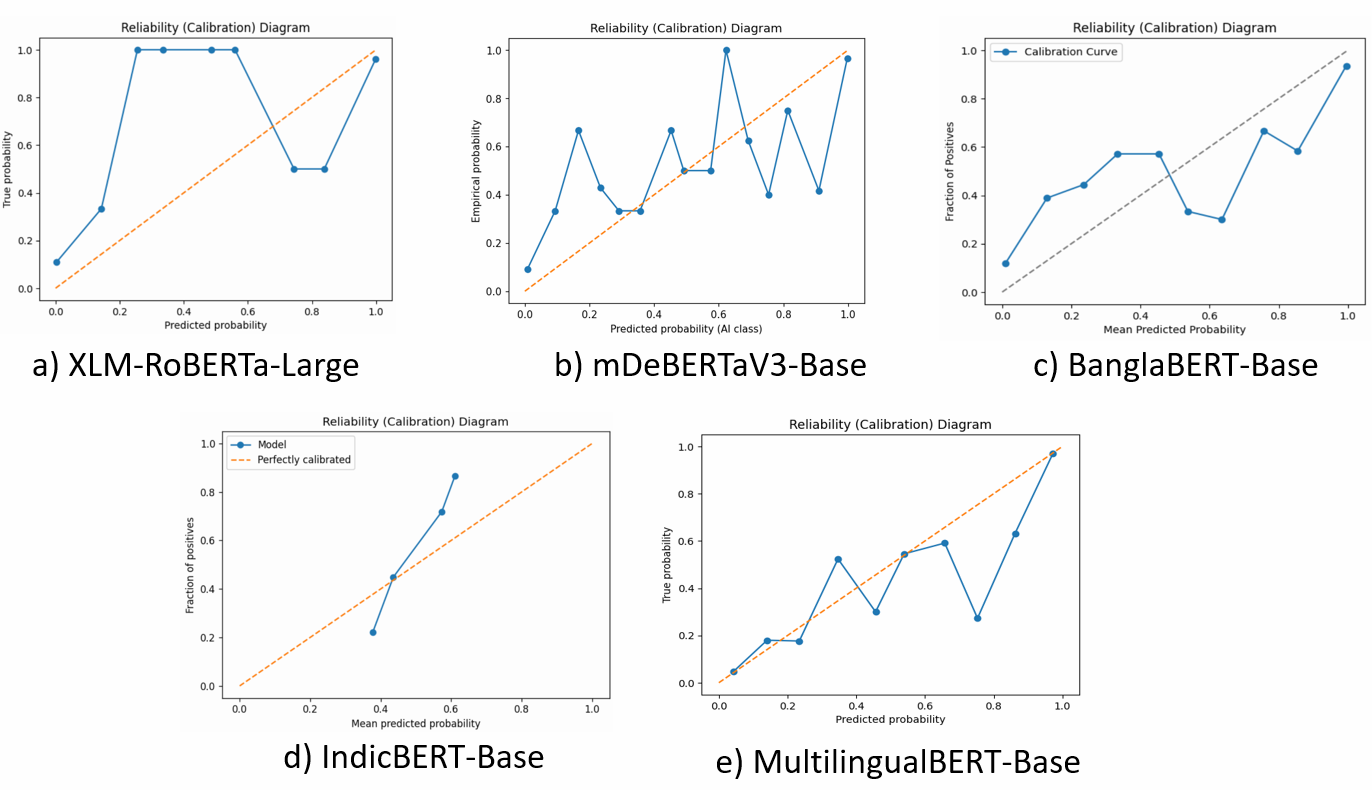}
    \caption{Reliability (Calibration) curve of the fine-tuned classification models.}
    \label{fig:gr5}
\end{figure}

The Reliability (Calibration) curves in Figure \ref{fig:gr5} compare how predicted probabilities align with true probabilities across five models. XLM-RoBERTa-Large shows significant overconfidence at low to mid probabilities and a sharp deviation before converging near high probabilities, indicating poor calibration despite good discrimination. mDeBERTaV3-Base fluctuates heavily around the diagonal, suggesting unstable probability estimates but less systematic bias. BanglaBERT-Base follows the diagonal more closely, though it slightly underestimates mid-range probabilities and overestimates at the high end, reflecting moderate calibration. IndicBERT-Base exhibits a smoother curve with consistent underconfidence at lower probabilities, indicating better reliability but less decisive predictions. MultilingualBERT-Base remains closer to the diagonal than XLM-RoBERTa-Large, with mild oscillations, reflecting relatively balanced calibration but some local misalignments.

\subsection{Comparison With Existing Work}
\cite{11022427} introduced the BanglaTextDistinguish dataset for detecting human-generated and AI-paraphrased Bengali text, showcasing results from several prominent models. We compare our results with theirs. Table \ref{tab:baseline_model_metrics} presents this comparison.

\begin{table}[!ht]
\centering
\caption{\label{tab:baseline_model_metrics} Performance comparison of baseline models for Bengali AI-paraphrased text detection. Accuracy, Precision, Recall and F1 Score values are presented in percentage.}
\begin{tabular}{|l|c|c|c|c|}
\hline
\textbf{Model Name} & \textbf{Acc} & \textbf{Prec} & \textbf{Rec} & \textbf{F1} \\
\hline
\makecell{SVM} & 76.99 & 80.59 & 72.27 & 76.20 \\
\makecell{RNN} & 72.36 & 44.57 & 73.80 & 73.33 \\
\makecell{BiLSTM} & 82.08 & 83.19 & 81.98 & 82.58 \\
\makecell{BiLSTM-SVM} & 82.83 & 82.85 & 82.83 & 82.84 \\
\makecell{\textbf{XLM-RoBERTa-Large} (Ours)} & \textbf{91.50} & \textbf{95.84} & \textbf{86.75} & \textbf{91.07} \\
\hline
\end{tabular}
\end{table}

Among the baseline models evaluated on the BanglaTextDistinguish dataset, XLM-RoBERTa-Large achieved the strongest performance with an accuracy of 91.50\%, precision of 95.84\% and an F1 score of 91.07\%, indicating highly reliable detection of AI-paraphrased text with minimal false positives. These results highlight the superiority of transformer-based approaches such as XLM-RoBERTa-Large over traditional and recurrent neural models for Bengali AI-generated text detection. Table \ref{tab:baseline_model_metrics} presents the detailed comparison.

\section{Conclusion and Future Works}
We present the first systematic study on detecting AI-generated Bengali text using transformer-based models. Zero-shot evaluation of XLM-RoBERTa-Large, mDeBERTaV3-Base, BanglaBERT-Base, IndicBERT-Base and MultilingualBERT-Base shows near-chance performance, while fine-tuning raises accuracy and F1 scores to around 91\% for XLM-RoBERTa, mDeBERTa and MultilingualBERT, with IndicBERT performing notably worse. Comparisons with BiLSTM-SVM and other baselines confirm the superiority of fine-tuned transformers on the BanglaTextDistinguish dataset. Future work should expand the dataset, explore cross-lingual transfer, improve robustness to unseen generation methods, develop lightweight models for real-time use and integrate linguistic or semantic cues to enhance detection reliability.

\bibliographystyle{ieeetr} 

\addcontentsline{toc}{chapter}{References}
\bibliography{ref}

\end{document}